%% file: iclr2025_conference.tex
\documentclass{article} 
\usepackage{iclr2025_conference,times}

\input{math_commands.tex}

\usepackage{hyperref}
\usepackage{url}
\usepackage{graphicx}
\usepackage{booktabs}
\usepackage{tabularx}

\usepackage{booktabs}
\usepackage{multirow}
\usepackage{makecell}
\usepackage{graphicx}

\usepackage{tabularx}
\usepackage{array}
\usepackage{subcaption}
\usepackage{algorithm}
\usepackage{algpseudocode}
\usepackage{placeins}
\usepackage{float}
\usepackage{fontawesome5}
\usepackage[table]{xcolor}

\usepackage{geometry}
\usepackage{fancyhdr}

\definecolor{citeblue}{RGB}{0,0,192}

\hypersetup{
    colorlinks=true,
    citecolor=citeblue,
    linkcolor=citeblue,
    urlcolor=citeblue
}

\renewcommand{\headrulewidth}{0pt}
\renewcommand{\footrulewidth}{0pt}
\fancypagestyle{plain}{
    \fancyhf{} 
    \renewcommand{\headrulewidth}{0pt}
    \renewcommand{\footrulewidth}{0pt}
    \fancyfoot[C]{\hspace{2cm}\thepage} 
}

\usepackage[many]{tcolorbox}
  \usepackage{enumitem}
  \usepackage{helvet}
  \usepackage{xcolor}

  \definecolor{PromptRule}{HTML}{5B6F8F}
  \definecolor{PromptBack}{HTML}{F7F9FC}
  \definecolor{PromptTitle}{HTML}{26364D}
  \definecolor{PromptCodeBack}{HTML}{F1F4F8}
  \definecolor{PromptCodeFrame}{HTML}{D8DEE8}
  \definecolor{headerblue}{RGB}{46, 107, 180}
  \definecolor{acapurple}{RGB}{155, 125, 195}

  \newtcolorbox{promptpanel}[1]{
      enhanced,
      breakable,
      colback=PromptBack,
      colframe=PromptRule,
      coltitle=PromptTitle,
      fonttitle=\bfseries,
      title=#1,
      attach boxed title to top left={xshift=2.5mm,yshift=-1.7mm},
      boxed title style={
          colback=PromptBack,
          colframe=PromptBack,
          boxrule=0pt,
          left=0pt,
          right=0pt,
          top=0pt,
          bottom=0pt
      },
      arc=0.6mm,
      boxrule=0.35pt,
      leftrule=1.2pt,
      top=3.8mm, bottom=2.2mm, left=3mm, right=3mm,
      before=\vspace{0.55em},
      after=\vspace{0.55em},
      fontupper=\small
  }

  \newtcolorbox{innercode}{
      enhanced,
      breakable,
      colback=PromptCodeBack,
      colframe=PromptCodeFrame,
      arc=0.5mm,
      boxrule=0.3pt,
      fontupper=\ttfamily\footnotesize,
      top=1.2mm, bottom=1.2mm, left=2mm, right=2mm
  }

\title{ECHO: Prune to Act, Trace to Learn \\ with Selective Turn Memory in Agentic RL \\[-0.3em]}

\author{
\vspace{-0.5cm}
\hspace*{-0.78em}
\begin{tabular}{@{}l@{}}
\textbf{Zijun Xie}$^{1,2 \ast \dagger}$
\quad
\textbf{Binbin Zheng}$^{2,3 \ast \dagger}$
\quad
\textbf{Enlei Gong}$^{2 \ast}$ 
\quad
\textbf{Jihua Liu}$^{2}$
\quad
\textbf{Yuyang You}$^{1}$\\[0.6ex]
\textbf{Lingfeng Liu}$^{1}$
\quad
\textbf{Jiayao Tang}$^{1}$
\quad
\textbf{Guanqun Zhao}$^{2}$
\quad
\textbf{Xiaoliang Fu}
\quad
\textbf{Aoqi Hu}$^{2}$
\quad
\textbf{Zeyu Chen}$^{2 \ddagger}$\\[1.4ex]
\normalfont $^{1}$School of Mathematical Sciences, Peking University
\quad
$^{2}$Baidu Inc. \\[0.8ex]
\normalfont $^{3}$University of Science and Technology of China \\[0.8ex]
\normalfont  \; \texttt{xiezijun@baidu.com}
\quad
{\small \codelink}
\end{tabular}
}

\newcommand{\codelink}{\href{https://github.com/xiezijun714-lang/Echo}{\faGithub\ \texttt{GitHub:xiezijun714-lang/Echo}}}

\iclrfinalcopy
\begin{document}

\maketitle

\makeatletter
\def\@oddhead{}
\def\@evenhead{}
\makeatother

\begin{abstract}
Long-horizon language agents must repeatedly interact with tools, accumulate evidence, and make decisions under bounded context windows. Context-management methods make such rollouts feasible by simplifying past
interactions through deletion, folding, or memory editing. However, when useful
history is collapsed into compressed states, the reconstructed context may no
longer reveal which earlier observations support a successful final answer. This creates a mismatch between bounded-context acting and outcome-based reinforcement learning: the policy acts on reconstructed context, while the learner lacks source-level provenance for assigning credit to the evidence that mattered. We propose ECHO, a selective turn-memory framework for traceable context reconstruction in agentic RL. ECHO compresses each completed environment turn into a compact source-indexed memory record, reconstructs bounded policy contexts by selecting useful records, and reuses the selected source indices to route positive outcome credit to the final trajectory segment, reused evidence turns, memory findings, and memory-selection actions. On BrowseComp-Plus, ECHO reaches 43.4\% held-out accuracy, outperforming GRPO at 28.9\% and the rolling-summary baseline SUPO at 36.1\%, while using fewer turns and lower trajectory volume than SUPO. The trained policy also improves zero-shot generalization across multi-objective QA, code generation, and deep information-seeking benchmarks on both dense and MoE backbones.
\end{abstract}

\begin{figure}[htbp]
    \centering
    \includegraphics[width=\linewidth]{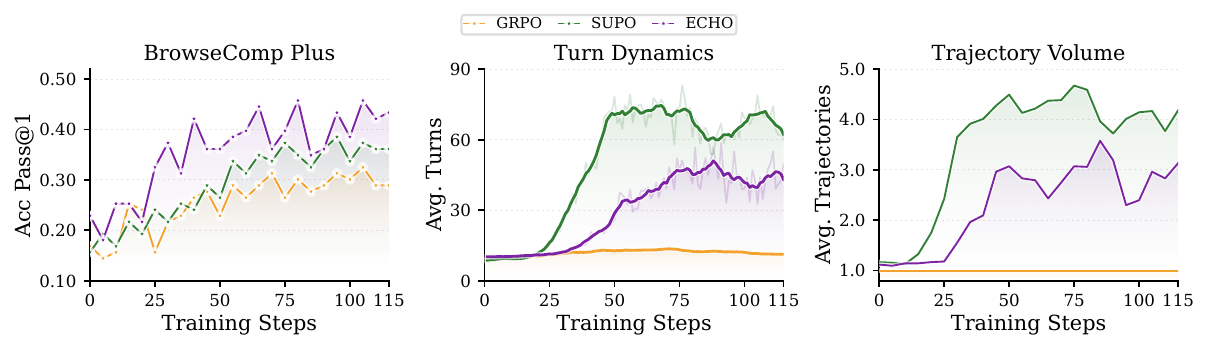}
    \caption{Held-out accuracy, tool-use turns per rollout, and trajectory volume over training on BrowseComp-Plus with the Qwen3-32B-Instruct backbone for \textsc{ECHO} (purple), GRPO (orange), and SUPO (green). \textsc{ECHO} traces the upper-left frontier: rising accuracy without the turn and volume growth seen for SUPO.}
    \label{fig:main-experiment}
\end{figure}

\section{Introduction}

Large language models (LLMs) are increasingly deployed as multi-turn agents
that interleave reasoning, tool invocation, and environment feedback
\citep{yao2023react, schick2023toolformer}. Reinforcement learning (RL) from
verifiable final outcomes has become a central recipe for improving such agents
in search, coding, function calling, and deep-research settings
\citep{jin2025searchr1, qian2025toolrl, li2025torl, zheng2025deepresearcher}.
As interaction horizons grow, however, history management becomes a bottleneck
for both acting and learning. The policy must retain useful observations within
a bounded context window, while the learner must decide which earlier decisions
should receive credit from a sparse final outcome.

Context-management methods make long rollouts feasible by simplifying past
interactions before the next decision. From the perspective of reconstructed
history, these methods differ in what form the completed history becomes:
some delete or omit distant turns, some edit an explicit memory state, and
others fold long prefixes into compact summaries or internal states. The
triggering mechanism may be rule-based or action-based, but this choice is
orthogonal to the reconstructed form of history. For outcome-based RL, the
central question is not only whether the shortened context is useful for
acting, but also whether it remains traceable to the original environment
turns for learning.

This provenance gap affects credit assignment in outcome-based agentic RL.
With final-outcome rewards, standard multi-turn RL often applies a
trajectory-level advantage to many generated action spans
\citep{schulman2017ppo, shao2024deepseekmath}. When the reconstructed context
is not source-addressable, such dense assignment can reinforce redundant
searches, incidental reasoning, or context-management behavior together with
genuinely useful evidence-gathering steps. The issue is therefore not only
whether context management enables longer bounded-context rollouts, but
whether the reconstructed context exposes the source turns that the learner
should credit. This suggests that a context manager should prune history for
acting while preserving source-level provenance for learning.

In this paper, we propose \textsc{ECHO}, a selective turn-memory framework for
traceable context reconstruction in agentic RL. Instead of collapsing distant
history into a single summary state, \textsc{ECHO} compresses each completed
tool-use turn into a compact memory record with a source-turn pointer. When the
context budget becomes binding, the policy selects useful memory records and
reconstructs the next bounded context from these selected records together with
recent interactions. The agent therefore acts with compact context, while the
selected memories remain explicitly linked to their original environment turns.

The same source-indexed reconstruction trace is then reused for learning.
Standard trajectory-level updates attach the group-relative advantage, whether
positive or negative, to all generated tokens in a rollout. In contrast,
\textsc{ECHO} first keeps only the positive part of the trajectory advantage,
and then applies it only to traceable credit tokens: the final trajectory segment,
selected historical source turns, their memory findings, and the
memory-selection actions that constructed the reconstructed context. Tokens
outside this trace receive no positive outcome credit. In this way,
\textsc{ECHO} changes credit assignment from dense trajectory-wide supervision
to provenance-guided positive credit routing, aligning context pruning for
acting with outcome-credit routing for learning through a single
source-indexed trace. Together, these design choices make context
reconstruction not only a way to fit long rollouts into a bounded context, but
also a learning interface for source-aware credit assignment.

Our main contributions are as follows:
\begin{itemize}
    \item We formulate context-managed multi-turn RL with a unified reconstruction
    interface that links history transformation to source traceability. This view
    shows why context management can make long rollouts feasible while still
    removing the provenance needed for outcome-credit assignment.
    
    \item We introduce \textsc{ECHO}, a turn-memory method that stores
    each turn as a source-indexed memory record and reconstructs
    bounded policy contexts through learned memory selection rather than
    collapsed-history folding.

    \item We propose provenance-guided token-level credit assignment, which
    reuses source indices to route positive outcome advantage to the final
    trajectory segment, historical evidence turns, memory findings, and
    memory-selection actions.

    \item Experiments on BrowseComp-Plus \citep{chen2025browsecompplus} and
    diverse zero-shot benchmarks show that \textsc{ECHO} improves accuracy while
    reducing turn proliferation and trajectory volume relative to the closest
    collapsed-history baseline, with consistent gains across dense and MoE
    backbones.
\end{itemize}

\section{Related Work} 
\subsection{RL for Long-Horizon Language Agents}

Language agents are commonly formulated as multi-turn decision makers that
interleave reasoning, tool invocation, and environment feedback
\citep{yao2023react}. Recent RL methods improve such agents with
outcome-based or verifiable rewards in search, function calling, coding, and
deep-research environments
\citep{schulman2017ppo, shao2024deepseekmath, guo2025deepseekr1,
jin2025searchr1, song2025r1searcher, li2025torl, qian2025toolrl,
zheng2025deepresearcher, du2025codegym}. These settings make context
management necessary during rollout, because useful observations may span
many tool-use turns and exceed the policy context budget. However, most
agentic RL objectives treat the reconstructed policy context as given, and do
not specify whether the information reused by the policy remains traceable to
the original environment turns. \textsc{ECHO} addresses this missing interface
between long-horizon rollout construction and learning by making reconstructed
context source-indexed.

\subsection{Context Management and Agent Memory}

Prior work addresses long-context limitations through compression,
summarization, retrieval, and memory modules
\citep{li2023compressing, li2024promptcompression, wang2024incontextformer,
xu2024concise, shen2025qwenlongcprs, yang2025pencil,
packer2023memgpt, zhong2024memorybank, chhikara2025mem0, xu2025amem}.
These methods improve bounded-context acting by shortening or externalizing
past interactions, but they usually do not define how final outcome rewards
should be linked back to the original turns that supplied useful evidence.

Recent agentic RL methods make context management trainable. Some replace the
distant prefix with updated summaries or compact states
\citep{wu2025resum, lu2025supo, yu2026memagent, zhou2025mem1}, while others
expose memory or context editing as actions
\citep{yan2025memoryr1, zhang2025memory, ning2026agentomit,
sun2025contextfolding}. Under the history-transformation view in
Table~\ref{tab:context_manager}, these methods differ in whether history is
deleted, edited, folded, or kept source-indexed. \textsc{ECHO} differs by
storing each completed turn as a source-indexed memory record and reusing the
selected records both for context reconstruction and outcome-credit routing.
This taxonomy motivates our empirical focus on SUPO-style rolling
summarization: it is the closest collapsed-history RL counterpart to
\textsc{ECHO} under the same agent-loop reconstruction setting, isolating the
difference between collapsed folding and source-indexed selection.

\subsection{Credit Assignment in Agentic RL}

Credit assignment is challenging when rewards are sparse and observed only
after extended interactions \citep{williams1992simple, sutton2018reinforcement}.
Many outcome-supervised RL methods reuse a trajectory- or response-level
advantage across generated action tokens or spans
\citep{schulman2017ppo, shao2024deepseekmath, yu2025dapo}. This dense
assignment is simple, but noisy in long-horizon tool-use tasks where only a
small subset of searches, observations, and intermediate decisions supports
the final answer.

Recent methods refine credit along temporal, hindsight, milestone,
hierarchical, or uncertainty-based axes
\citep{zeng2025turnlevel, tan2026hcapo, wang2026beacon, peng2026hiper,
zhang2026creditassignment, zhao2026aem}. These approaches improve how rewards
are distributed over time or policy structure, but are mostly independent of
how the agent's effective context was reconstructed. \textsc{ECHO} is
complementary: rather than estimating a new temporal advantage, it uses the
source indices exposed by context reconstruction to define a
provenance-guided token mask for positive outcome credit.

\section{Preliminaries}
\label{sec:prelim}

\subsection{Multi-Turn GRPO for Agentic RL}
\label{sec31}

Let $x$ denote an initial task prompt. At interaction step $t$, the agent
conditions on a policy context $c_t$ and samples an action
$a_t \sim \pi_\theta(\cdot \mid c_t)$. The action may contain reasoning
tokens, a tool invocation, an internal context-management operation, or a
final answer. If a tool is invoked, the environment returns an observation
$o_t$; otherwise $o_t$ may be null. The resulting multi-turn rollout is denoted
by
\begin{equation}
\tau=(a_1,o_1,\ldots,a_T,o_T),
\end{equation}
and receives a sparse final reward $R(\tau)$.

For each prompt, GRPO \citep{shao2024deepseekmath} samples a group of $N$
complete trajectories $\{\tau^{(1)},\ldots,\tau^{(N)}\}$. Let
$r^{(n)}=R(\tau^{(n)})$, and let $\bar r$ and $s_r$ be the mean and standard
deviation of $\{r^{(i)}\}_{i=1}^{N}$. The group-relative advantage is
\begin{equation}
A^{(n)}=\frac{r^{(n)}-\bar r}{s_r+\epsilon}.
\label{eq:grpo-advantage}
\end{equation}

We use the following simplified policy-gradient form:
\begin{equation}
\mathcal{J}_{\mathrm{MT}}(\theta)
=
\mathbb{E}_{x\sim\mathcal D}
\left[
\frac{1}{N}
\sum_{n=1}^{N}
\sum_{t=1}^{T_n}
A^{(n)}
\log \pi_\theta(a_t^{(n)} \mid c_t^{(n)})
\right].
\label{eq:mt-grpo}
\end{equation}
This expression suppresses implementation-specific surrogate terms, such as
old-policy ratios, clipping or gating, token normalization, and optional KL
penalties. Its purpose is to make explicit that a trajectory-level advantage is
attached to generated actions conditioned on the policy contexts actually
provided to the model. In token-level implementations, each action $a_t$ is
expanded into generated tokens, and the corresponding token log probabilities
share the same trajectory-level advantage.

Here $c_t$ denotes the bounded policy context used for the current model call.
Standard multi-turn RL objectives typically treat this context as given. The
next subsection makes the context-management mechanism explicit by separating
the complete environment-side history from the reconstructed policy context.

\begin{table*}[t]
\centering
\small
\setlength{\tabcolsep}{5pt}
\renewcommand{\arraystretch}{1.25}
\caption{Context-managed rollout strategies under the $(\mathcal{M},\Phi)$
interface. We organize methods by how they transform completed history.
Superscripts on examples indicate how reconstruction is triggered:
$\mathrm{R}$ denotes rule-triggered reconstruction, and $\mathrm{A}$ denotes
action-triggered reconstruction. Source trace indicates whether the
reconstructed context remains addressable at the level of original environment
turns.}
\label{tab:context_manager}
\begin{tabular}{@{} l l l l l @{}}
\toprule
\textbf{History Transformation}
& \textbf{Managed State}
& \textbf{Reconstruction Form}
& \textbf{Source Trace}
& \textbf{Examples} \\
\midrule
Append-only
& $H_j^{\mathrm{pre}}$
& $H_j^{\mathrm{pre}}\oplus H_{j,t}^{\mathrm{loc}}$
& Explicit
& Vanilla prompting \\

Deletion / pruning
& $C_j\subseteq H_j^{\mathrm{pre}}$
& $\mathrm{Render}(C_j,H_{j,t}^{\mathrm{loc}};B)$
& Partial / lost
& Sliding window$^{\mathrm{R}}$, Agent-Omit$^{\mathrm{A}}$ \\

Edited memory state
& $B_j$
& $\mathrm{Render}(B_j,H_{j,t}^{\mathrm{loc}};B)$
& Action-traced
& MemAct$^{\mathrm{A}}$, Memory-R1$^{\mathrm{A}}$ \\

Collapsed folding
& $z_j=C(z_{j-1},\sigma_{j-1})$
& $\mathrm{Render}(z_j)\oplus H_{j,t}^{\mathrm{loc}}$
& Collapsed
& SUPO$^{\mathrm{R}}$, FoldGRPO$^{\mathrm{A}}$ \\

\midrule
Selective turn memory
& $\mathcal{E}_j=\{e_i\}_{i\le K_{j-1}}$
& $\mathrm{Render}(\mathcal{E}_j[\hat I_j])\oplus H_{j,t}^{\mathrm{loc}}$
& \textbf{Source-indexed}
& \textbf{\textsc{ECHO}$^{\mathrm{R}}$} \\
\bottomrule
\end{tabular}
\end{table*}

\subsection{Context-Managed Multi-Turn Rollouts}
\label{sec32}

In long-horizon tool-use tasks, the complete environment-side history may
exceed the context budget used during rollout and training. We write one
completed interaction turn as $u_t=(a_t,o_t)$, so that the full rollout can be
equivalently written as $\tau=(u_1,\ldots,u_T)$. Let $B$ be the maximum policy
context length. A context-managed rollout must construct a policy context
$c_t$ satisfying $|c_t|\le B$ at every interaction step.

For clarity, we first describe one trajectory and omit the trajectory
superscript. We partition the rollout into bounded-context segments:
\begin{equation}
0=K_0<K_1<\cdots<K_J=T.
\end{equation}
The boundaries $\{K_j\}$ may be triggered by rule-based controllers, such as
length or turn-count conditions, or by policy-generated context-management
actions. The turns
in segment $j$ are indexed by
$\mathcal{T}_j=\{K_{j-1}+1,\ldots,K_j\}$. For segment $j$, the completed
prefix and the local within-segment history before turn $t$ are
\begin{equation}
H_j^{\mathrm{pre}}
=(u_1,\ldots,u_{K_{j-1}}),\quad
H_{j,t}^{\mathrm{loc}}=(u_{K_{j-1}+1},\ldots,u_{t-1}),
\quad
K_{j-1}<t\le K_j .
\end{equation}

A context manager maps the completed prefix into a managed state, and a
reconstruction function renders the bounded policy context:
\begin{equation}
z_j = \mathcal{M}(H_j^{\mathrm{pre}}), \quad
c_{j,t} = x \oplus
\Phi\!\left(z_j, H_{j,t}^{\mathrm{loc}}; B\right).
\label{eq:context-reconstruction}
\end{equation}
Here $z_j$ may be retained history, a summary, a compressed state, an edited
memory, or a set of memory units. The function $\Phi$ combines $z_j$ with the
local history under the budget $B$, and $\oplus$ denotes prompt concatenation,
or more generally the effective conditioning mechanism.

This interface separates two properties of a context manager: whether it keeps
the policy context short enough for acting, and whether the reconstructed
information remains source-addressable for learning. Traceable methods may
expose metadata such as source indices of retained or selected turns, which
can later be used for credit assignment.

With context reconstruction, the reward and group-relative advantage remain
unchanged. What changes is only the conditioning path from completed history
to policy input:
\[
H_j^{\mathrm{pre}}
\xrightarrow{\mathcal{M}}
z_j
\xrightarrow{\Phi(\cdot, H_{j,t}^{\mathrm{loc}};B)}
c_{j,t}.
\]
At the objective level, this simply replaces the original policy context
$c_t^{(n)}$ in Eq.~\ref{eq:mt-grpo} with the reconstructed segment context
$c_{j,t}^{(n)}$. For trajectory $n$, let $\mathcal{T}_j^{(n)}$ denote the turn
indices of segment $j$. Then
\begin{equation}
\begin{aligned}
\mathcal{J}_{\mathrm{CM}}(\theta)
=
\mathbb{E}_{x\sim\mathcal D}
\left[
\frac{1}{N}
\sum_{n=1}^{N}
\sum_{j=1}^{J_n}
\sum_{t\in\mathcal{T}_j^{(n)}}
A^{(n)}
\log \pi_\theta
\left(a_t^{(n)} \mid c_{j,t}^{(n)}\right)
\right].
\end{aligned}
\label{eq:cm-grpo}
\end{equation}
The segment index is only bookkeeping: the segments partition the same
rollout, and the learning signal remains the trajectory-level advantage
$A^{(n)}$. Thus, different context managers differ not by the reward estimator
above, but by their choices of $\mathcal{M}$ and $\Phi$, which determine the
reconstructed context and its source trace. The key question is whether this
trace remains addressable at the level of original environment turns, so that
outcome credit can be routed to the evidence that the policy actually reused.

\section{Method}
\label{methods}

\subsection{Why Traceability Matters}
\label{sec:why-trace}

The reconstruction interface in Eq.~\ref{eq:context-reconstruction} separates
two questions that are often conflated in long-horizon agentic RL: whether the
policy can act under a bounded context budget, and whether the bounded context
remains traceable to original environment turns. Table~\ref{tab:context_manager}
organizes context managers by how they transform completed history. Append-only
prompting preserves explicit trace but exceeds the context budget; deletion
and pruning remove turns; edited-memory methods maintain compact memory states;
and collapsed-folding methods replace long prefixes with summaries or compact
states. These transformations make long rollouts feasible, but differ in
whether reconstructed context remains source-addressable for learning.

This distinction matters because outcome-based RL assigns credit after the
final answer is verified. If the policy conditions on reconstructed context
whose source turns are no longer addressable, the learner cannot tell
which earlier observations supported the successful decision. A dense
trajectory-level update may then reinforce useful evidence-gathering turns
together with redundant searches, incidental reasoning, or summary-generation
tokens. Thus, context efficiency alone is not enough: the reconstruction
process should expose a trace of the evidence that was selected and reused.

\begin{figure}[t]
    \centering
    \includegraphics[width=\linewidth]{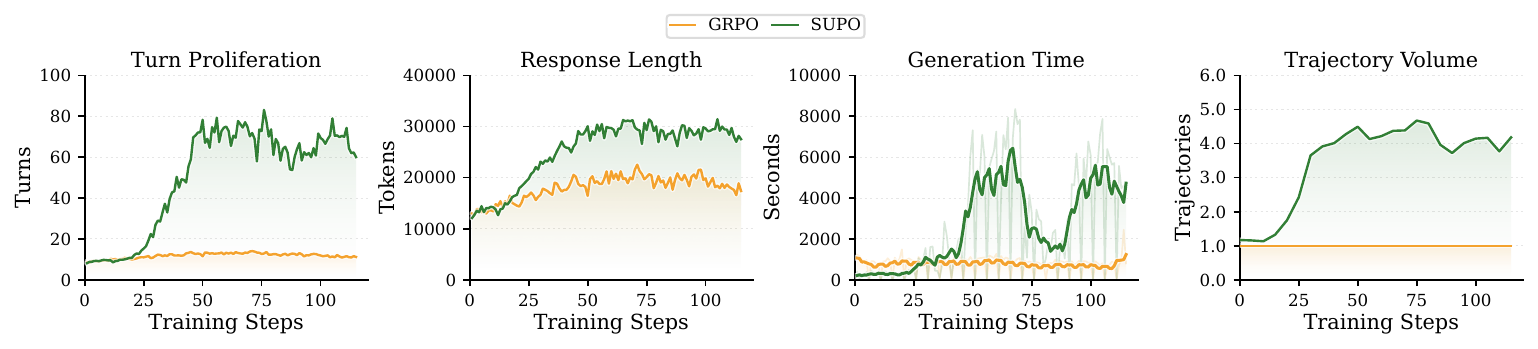}
    \caption{
    Training diagnostics on long-horizon search. Summarization-based context
management enables longer rollouts, but leads to turn proliferation,
longer responses, higher generation time, and increased trajectory volume.
    }
    \label{fig:motivation-diagnostics}
\end{figure}

Collapsed summarization illustrates the limitation of history compression
alone. By folding distant interactions into a compact summary state, it allows
the agent to continue beyond the raw-context limit. However, the resulting
context no longer exposes which original turns supplied the reused evidence.
Figure~\ref{fig:motivation-diagnostics} shows the practical consequence:
summarization-based context management extends rollouts, but also produces
more tool-use turns, longer responses, higher generation time, and larger
trajectory volume. This supports the concern above: making history shorter for
acting does not necessarily make learning easier. A context manager should
therefore not only compress history, but also preserve explicit source traces
that can support outcome-credit assignment.

\subsection{Selective Turn Memory}
\label{sec:selective-memory}

\textsc{ECHO} instantiates source-indexed reconstruction with selective turn
memory. The goal is to keep the active policy context compact without folding
the distant prefix into a single untraceable state. To do so, \textsc{ECHO}
separates local turn compression from global context reconstruction: each
completed turn is compressed independently into a source-indexed memory record,
and the next policy context is reconstructed by selecting useful records from
the resulting memory set.

\paragraph{Source-indexed memory.}
After each completed tool-use turn $u_i=(a_i,o_i)$, \textsc{ECHO} asks the
policy to write a compact local finding:
\begin{equation}
s_i \sim \pi_\theta(\cdot \mid c_i^{\mathrm{sum}}),
\qquad
m_i=\mathrm{Parse}(s_i).
\label{eq:turn-summary}
\end{equation}
Here $c_i^{\mathrm{sum}}$ contains the latest completed turn and a summary
instruction that asks the model to summarize only that turn. This finding is
generated by the same policy being trained, rather than by an external
summarizer or a separate memory model. The memory record is
\begin{equation}
e_i=(i,\alpha_i,m_i),
\qquad
\mathcal{E}_j=\{e_i\}_{i\le K_{j-1}} .
\label{eq:memory-record}
\end{equation}
The index $i$ is the source-turn pointer, $\alpha_i$ is a compact rendering of
the action or tool call, and $m_i$ is the parsed last-turn finding. Unlike a
rolling summary, $\mathcal{E}_j$ is a persistent, non-collapsing archive.
Selection changes only the active view rendered into the next policy context;
it does not delete unselected records. Thus, every memory remains individually
addressable at later reconstruction boundaries even after its raw observation
is no longer placed in the active context.

\paragraph{Memory selection.}
At a compression boundary before segment $j$, \textsc{ECHO} lets the policy
select historical memories that are useful for continuing the task. Let
$H_j^{\mathrm{bd}}$ denote the bounded local state available at the boundary.
The selection context is
\begin{equation}
\begin{aligned}
c_j^{\mathrm{sel}}
= x
&\oplus \Phi_{\mathrm{local}}(H_j^{\mathrm{bd}};B)
\oplus \mathrm{RenderList}(\mathcal{E}_j)
\oplus x_{\mathrm{sel}},
\end{aligned}
\end{equation}
and the policy generates a selection action
\begin{equation}
a_j^{\mathrm{sel}}
\sim \pi_\theta(\cdot \mid c_j^{\mathrm{sel}}).
\label{eq:selection-action}
\end{equation}
The instruction $x_{\mathrm{sel}}$ asks the model to output source indices of
reusable evidence, constraints, failed attempts, or useful next-step plans. A
repair operator converts the generated text into a valid bounded set:
\begin{equation}
\hat I_j
=
\rho_{B,S,K}
\left(
\mathrm{Parse}(a_j^{\mathrm{sel}}),
\mathcal{E}_j
\right).
\label{eq:repair-selection}
\end{equation}
The operator $\rho_{B,S,K}$ removes malformed or out-of-range indices,
deduplicates selections, keeps at most $S$ model-selected historical turns,
and merges them with the latest $K$ turns retained automatically. The rendered
view is then validated against the context budget; a rollout is marked overlong
if the bounded view still cannot be constructed. Thus, both memory writing and
memory selection are policy-generated and trained within the same RL loop.

\paragraph{Context reconstruction.}
The selected indices determine which memories are used for reconstruction:
\begin{equation}
\mathcal{E}_j[\hat I_j]
=
\{e_i\in \mathcal{E}_j \mid i\in \hat I_j\}.
\label{eq:selected-memory}
\end{equation}
For any turn $t$ in segment $j$, \textsc{ECHO} reconstructs the policy context
as
\begin{equation}
\begin{aligned}
c_{j,t}
=
x
&\oplus \mathrm{Render}(\mathcal{E}_j[\hat I_j])
\oplus \Phi_{\mathrm{local}}(H_{j,t}^{\mathrm{loc}};B),
\end{aligned}
\label{eq:echo-context}
\end{equation}
with $|c_{j,t}|\le B$. The agent acts on a compact context composed
of selected memories and recent interactions. At the same time, the selected
set $\hat I_j$ preserves a source-indexed reconstruction trace, which is
reused for credit routing.

\subsection{Traceable Credit Routing}
\label{sec:traceable-credit}

\begin{figure*}[t]
    \centering
    \includegraphics[width=0.98\textwidth]{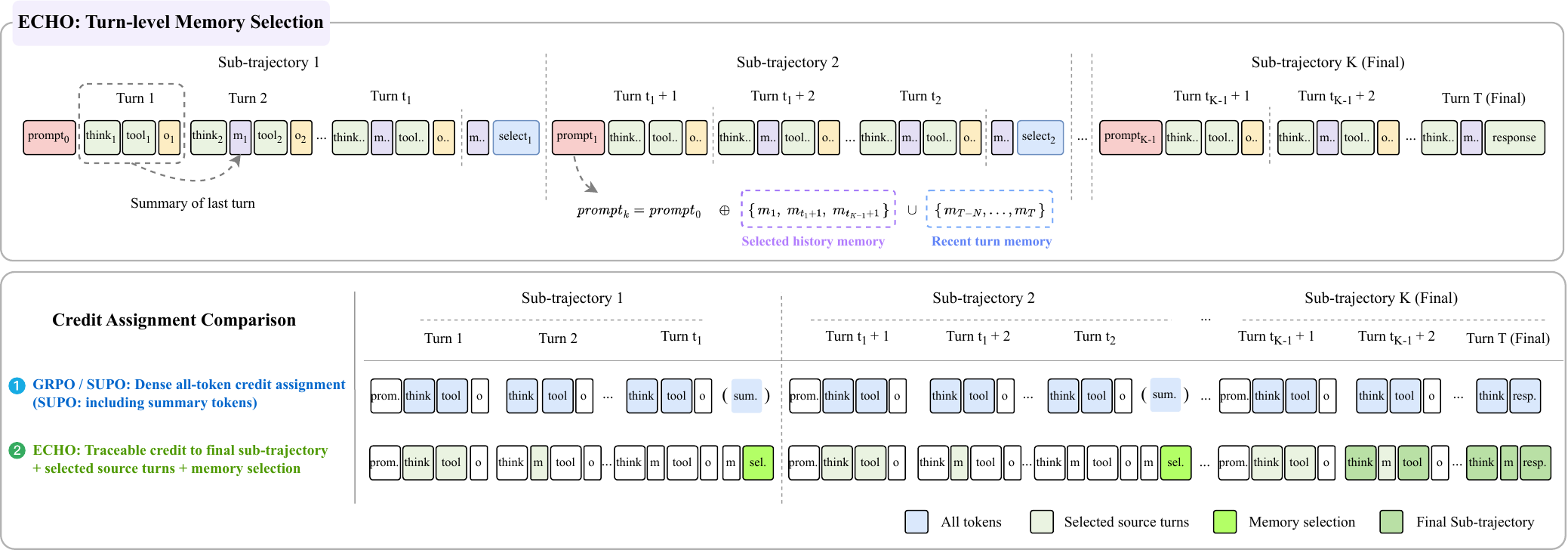}
    \caption{\textbf{Overview of ECHO.} ECHO stores completed turns as
    source-indexed memories, selects useful memories for bounded context
    reconstruction, and reuses the same source trace for credit assignment.}
    \label{fig:echo-overview}
\end{figure*}

\textsc{ECHO} uses a final-trace approximation for outcome-based learning: it
routes credit through the source turns selected into the final reconstructed
context, rather than recursively tracking all intermediate reconstruction
dependencies. This final reconstruction trace defines which generated tokens
are eligible for positive outcome credit.

For trajectory $n$, let $I_{\mathrm{src}}^{(n)}$ be the source turns selected
into the final reconstructed context under this final-trace approximation. Let
$j$ index saved trajectory segments and $q$ index a generated response-token
position within segment $j$. For each token, we record four indicators:
$g_{j,q}^{(n)}$ marks trainable policy tokens in the final saved segment,
$d_{j,q}^{(n)}$ gives the source turn of a normal assistant/action token,
$f_{j,q}^{(n)}$ gives the source turn of a generated memory-finding token, and
$b_{j,q}^{(n)}$ marks memory-selection tokens. \textsc{ECHO} defines the hard
credit mask
\begin{equation}
\mu_{j,q}^{(n)}
=
\mathbf{1}\big[
g_{j,q}^{(n)}=1
\vee d_{j,q}^{(n)}\in I_{\mathrm{src}}^{(n)}
\vee f_{j,q}^{(n)}\in I_{\mathrm{src}}^{(n)}
\vee b_{j,q}^{(n)}=1
\big].
\label{eq:credit-mask}
\end{equation}
The mask marks tokens that are eligible for positive credit: all trainable
tokens in the final trajectory segment, action tokens from selected source
turns, their memory-finding tokens, and generated memory-selection spans. The
final segment begins after the last context reconstruction and includes its
assistant reasoning and tool-call tokens together with the terminal response.

For binary verifier rewards, the selected trace is reliable only when the
rollout is better than the group baseline. If the rollout fails, selected turns
may be irrelevant, misleading, or useful but misused. We therefore route only
the positive part of the group-relative advantage:
\begin{equation}
A_+^{(n)}=\max(A^{(n)},0),
\qquad
\widetilde{A}_{j,q}^{(n)}=\mu_{j,q}^{(n)}A_+^{(n)} .
\label{eq:positive-credit}
\end{equation}
Because credit is multiplied by $A_+^{(n)}$, only positive-advantage rollouts
assign nonzero credit to the masked tokens. Memory-selection tokens are
included in the eligible set because they choose the memories that later
successful decisions rely on.
The per-trajectory traceable surrogate is
\begin{equation}
\mathcal{J}_{\mathrm{ECHO}}^{(n)}(\theta)
=
\sum_j\sum_q
\widetilde A_{j,q}^{(n)}
\log \pi_\theta
\left(
y_{j,q}^{(n)}
\mid
c_j^{(n)},y_{j,<q}^{(n)}
\right).
\label{eq:echo-objective}
\end{equation}
This surrogate has the same simplified policy-gradient interpretation as
Eq.~\ref{eq:mt-grpo}, but replaces dense trajectory-wide weighting with the
trace-masked positive weight $\widetilde{A}_{j,q}^{(n)}$. Thus, \textsc{ECHO} turns
source-indexed reconstruction into a token-level credit route.

\section{Experiment}

\begin{figure}[t]
    \centering
    \includegraphics[width=\linewidth]{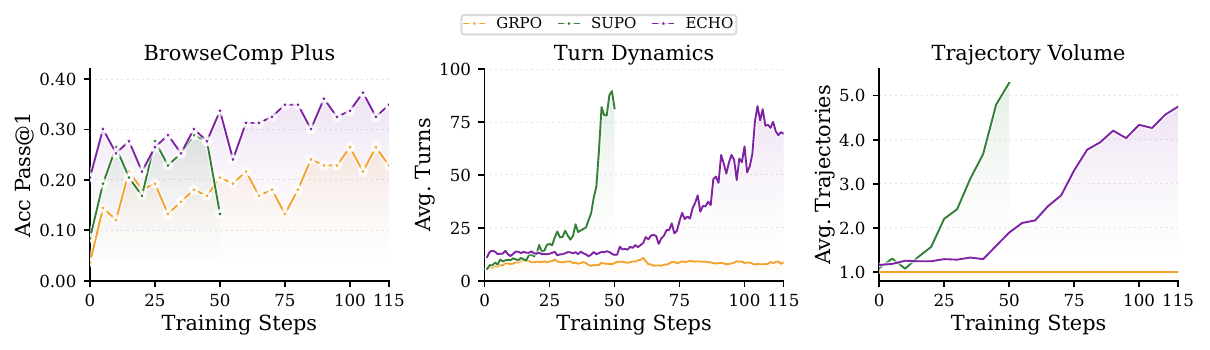}
    \caption{
    Training dynamics on BrowseComp-Plus with the Qwen3-30B-A3B-Instruct MoE
    backbone. The panels report held-out accuracy, average tool-use turns, and
    trajectory volume; \textsc{ECHO} improves accuracy without SUPO's turn and
    volume growth.
    }
    \label{fig:moe-experiment}
\end{figure}

\subsection{Experimental Setup}
\label{sec:experimental-setup}

We use SUPO-style rolling summarization as the main context-managed RL baseline
because it is the closest collapsed-history counterpart to \textsc{ECHO}. Both
methods perform context reconstruction inside the agent loop and train the
policy to generate the text used for context management, rather than relying on
an external retriever or a separate memory model. The key difference is the
form of reconstructed history: SUPO folds distant interactions into a collapsed
summary state, whereas \textsc{ECHO} reconstructs context from selected
source-indexed turn memories and reuses the selected source trace for credit
routing.

We evaluate on BrowseComp-Plus, a long-horizon tool-use QA benchmark, and
compare GRPO \citep{shao2024deepseekmath}, SUPO-style rolling summarization
\citep{lu2025supo}, and \textsc{ECHO}. All methods use the same search/open-page
tool environment, verifier, 8 rollouts per prompt, and 32k-token working context
budget. For context-managed methods, we allow at most 5 reconstruction rounds,
which gives an effective interaction budget of up to 192k tokens while keeping
each policy call within the 32k working context. \textsc{ECHO} automatically
retains the latest 3 turns and selects up to 8 additional historical source
turns at each reconstruction boundary. We report held-out pass@1 and
rollout-efficiency metrics in the main text; dataset statistics, retrieval
settings, and zero-shot benchmark details are provided in the appendix.

\subsection{Main Results}
\label{sec:main-results}
Figure~\ref{fig:main-experiment} shows that \textsc{ECHO} reaches 43.4\% held-out
accuracy on the dense backbone, outperforming GRPO (28.9\%) and SUPO (36.1\%).
GRPO remains short but plateaus early, indicating that raw bounded-context
training limits the amount of evidence the agent can gather. SUPO improves
accuracy by extending rollouts, but this comes with substantial turn
proliferation: its final turn count reaches 62.5, with an 85.5\% trajectory
split rate and 4.18 trajectories per rollout. \textsc{ECHO} surpasses SUPO while
ending with fewer turns (45.3), a lower split rate (57.8\%), and fewer
trajectories per rollout (3.13). These results suggest that \textsc{ECHO}'s
source-indexed selection and traceable credit routing help the agent reuse
useful evidence more selectively, yielding more effective search paths under
the same bounded-context setting.

Figure~\ref{fig:moe-experiment} shows the same ranking with the sparse
Qwen3-30B-A3B-Instruct MoE backbone. GRPO reaches only 22.9\% accuracy. SUPO
initially benefits from rolling summaries, but its turn count and trajectory
volume grow rapidly and accuracy drops to 13.3\% by step 50. In contrast,
\textsc{ECHO} remains stable and finishes around 35.0\%. This suggests that
source-indexed selection and traceable credit routing are not specific to the
dense backbone.

\begin{figure*}[t]
    \centering
    \begin{subfigure}[t]{0.98\textwidth}
        \centering
        \includegraphics[width=\linewidth]{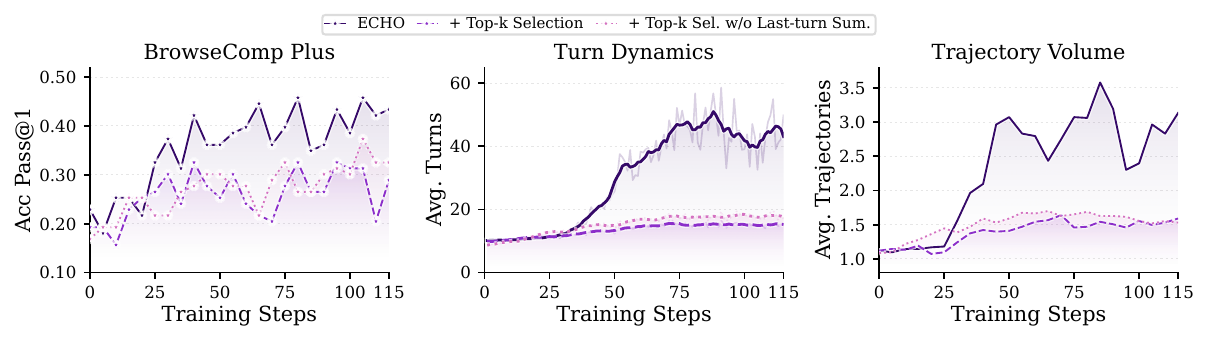}
        \caption{Memory component ablation.}
        \label{fig:component-ablation}
    \end{subfigure}
    \hfill
    \begin{subfigure}[t]{0.98\textwidth}
        \centering
        \includegraphics[width=\linewidth]{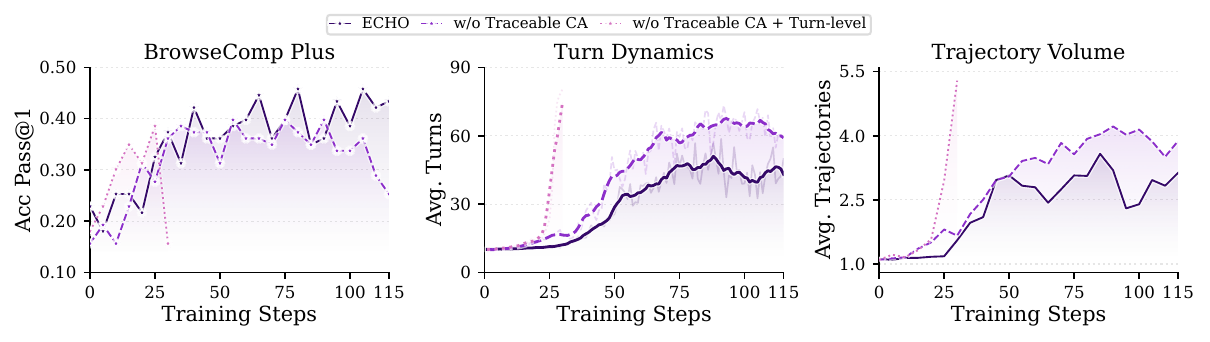}
        \caption{Credit assignment ablation.}
        \label{fig:credit-ablation}
    \end{subfigure}
    \caption{
    Ablations on memory components and credit assignment. Learned source
    selection is strongest, compact findings are sufficient, and removing
    traceable credit routing reduces accuracy and stability.
    }
    \label{fig:ablation-comparison}
\end{figure*}

\subsection{Ablation Study}
\paragraph{Memory ablations.}
Figure~\ref{fig:component-ablation} studies how the memory component should be
constructed and retrieved. Replacing learned source selection with static
semantic top-$k$ retrieval keeps the reconstructed context compact, but yields
lower accuracy than full \textsc{ECHO}. This suggests that the benefit of
turn memory does not come only from storing compressed historical records; the
policy must also learn which records are useful for the current decision.
Replacing compact turn findings with full observations also brings no clear
gain, indicating that local findings are sufficient for preserving reusable
evidence while avoiding unnecessary context growth. Together, these results
support the design choice of storing compact source-indexed findings and
performing policy-driven selection at reconstruction time.

\paragraph{Credit-routing ablations.}
Figure~\ref{fig:credit-ablation} evaluates whether the selected source trace
should also guide credit assignment. Removing traceable credit assignment lowers
accuracy and leads to longer interaction paths with more turns and larger
trajectory volume. This suggests that source-indexed reconstruction alone is
not sufficient: if the final outcome advantage is densely assigned to all
generated tokens, the learner may still reinforce redundant searches,
incidental reasoning, and other tokens that merely co-occur with a successful
rollout. Adding turn-level importance sampling does not solve this issue; in
our training curves, it becomes unstable and collapses as turn count and
trajectory volume grow. In contrast, \textsc{ECHO} routes positive credit only
through the final trajectory segment, selected evidence turns, their memory findings, and
the selection actions that exposed those memories. This acts as a denoising
mechanism for outcome credit and helps compress successful search paths rather
than merely extending them.

\begin{table*}[t]
\centering
\caption{
Zero-shot generalization results across various benchmarks. \textbf{Bold}
denotes the best performance and \underline{underlined} denotes the second-best.
CA denotes credit assignment, Traceable CA denotes source-indexed credit routing,
and Turn-level IS denotes importance-sampling ratios aggregated at the turn
level.
}
\vspace{6pt}
\renewcommand{\arraystretch}{1.25}
\resizebox{\textwidth}{!}{
\begin{tabular}{l c c c c c c c c c c c}
\toprule
\multirow{2}{*}{\textbf{Method}}
& \multicolumn{5}{c}{\textbf{Multi-Objective QA}}
& \multicolumn{2}{c}{\textbf{Code Generation}}
& \multicolumn{3}{c}{\textbf{Deep Information Seeking}}
& \multirow{2}{*}{\textbf{Avg.}} \\
\cmidrule(lr){2-6}
\cmidrule(lr){7-8}
\cmidrule(lr){9-11}
& \textbf{2-obj.}
& \textbf{4-obj.}
& \textbf{8-obj.}
& \textbf{16-obj.}
& \textbf{Avg.}
& \textbf{CodeGym}
& \textbf{LoCoBench-Agent}
& \textbf{GAIA}
& \textbf{HLE}
& \textbf{Frames}
& \\
\midrule
\rowcolor{headerblue!25}
\multicolumn{12}{c}{\textit{\textbf{Backbone: Qwen3-32B-Instruct}}} \\
GRPO
& 38.6 & 39.8 & 35.8 & 29.0 & 35.8
& 32.8 & 67.7
& \underline{25.2} & 8.8 & 24.8 & 33.6\\
SUPO
& 40.9 & 36.4 & 36.4 & 34.7 & 37.1
& 35.4 & 68.1
& \underline{25.2} & 9.2 & 26.8 & 34.8\\
ECHO w/ Top-K retrieval
& \textbf{47.7} & 42.0 & \underline{39.2} & 27.8 & \underline{39.2}
& \underline{40.7} & 69.3
& \textbf{29.1} & \underline{10.6} & \underline{37.3} & \underline{38.2}\\
ECHO w/ Top-K retrieval \& w/o turn summary
& 40.9 & 44.3 & 35.8 & \underline{35.5} & 39.1
& 40.3 & 69.5
& 23.3 & 10.0 & 31.3 & 36.8\\
ECHO w/o traceable CA
& \underline{45.5} & 42.0 & \underline{39.2} & 22.7 & 37.4
& 38.1 & 68.2
& \underline{25.2} & 8.8 & 30.8 & 35.6\\
ECHO w/o Traceable CA \& Turn-level IS
& \underline{45.5} & \textbf{47.7} & 35.2 & 27.0 & 38.8
& 34.6 & \underline{70.1}
& 23.3 & 9.4 & 32.2 & 36.1\\
\rowcolor{acapurple!25}
ECHO
& \textbf{47.7} & \underline{45.5} & \textbf{41.5} & \textbf{36.1} & \textbf{42.7}
& \textbf{41.4} & \textbf{70.4}
& \textbf{29.1} & \textbf{11.4} & \textbf{39.1} & \textbf{40.2}\\
\midrule
\rowcolor{headerblue!25}
\multicolumn{12}{c}{\textit{\textbf{Backbone: Qwen3-30B-A3B-Instruct}}} \\
GRPO
& \underline{27.3} & 26.1 & \underline{27.3} & 16.2 & 24.2
& 20.3 & \underline{65.7}
& \underline{23.3} & 7.8 & \underline{19.1} & 25.9 \\
SUPO
& 25.0 & \underline{30.7} & \underline{27.3} & \underline{18.2} & \underline{25.3}
& \underline{27.3} & 65.1
& \underline{23.3} & \underline{8.0} & 17.0 & \underline{26.9} \\
\rowcolor{acapurple!25}
ECHO
& \textbf{34.1} & \textbf{36.4} & \textbf{30.1} & \textbf{18.8} & \textbf{29.9}
& \textbf{29.7} & \textbf{66.8}
& \textbf{24.3} & \textbf{9.2} & \textbf{25.0} & \textbf{30.5} \\
\bottomrule
\end{tabular}
}
\label{tab:main_results}
\end{table*}

\subsection{Training Efficiency}
\begin{figure}[t]
    \centering
    \includegraphics[width=\linewidth]{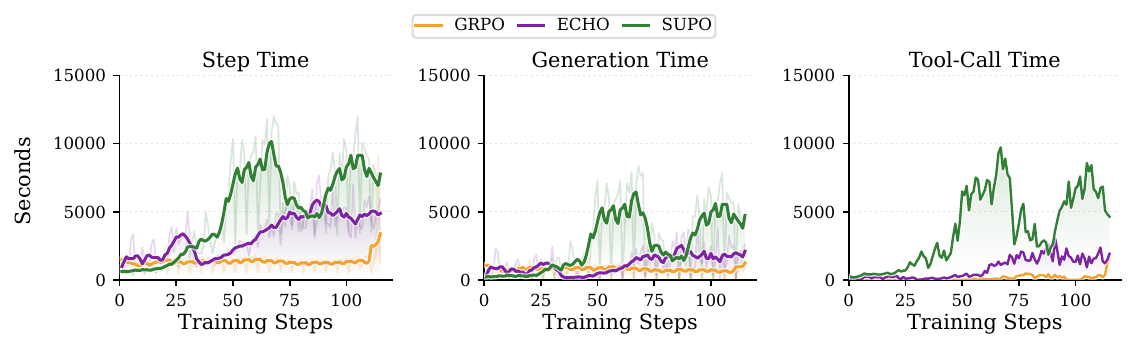}
    \caption{
    Training-time efficiency on BrowseComp-Plus. \textsc{ECHO} keeps step,
    generation, and tool-call time lower than SUPO.
    }
    \label{fig:training-efficiency}
\end{figure}

Figure~\ref{fig:training-efficiency} compares wall-clock cost during
dense-backbone training. \textsc{ECHO} introduces early overhead because each
completed tool turn writes a \texttt{sum\_last\_turn} finding and compression
boundaries add selection actions. As training proceeds, however, SUPO's rolling
summaries lead to rapidly growing interaction rounds, making tool calls the
dominant cost. By tying reconstruction to a bounded set of selected source
turns, \textsc{ECHO} limits this mid-to-late-stage turn proliferation and
achieves higher accuracy with fewer splits and lower tool-call-dominated
execution time.

\subsection{Zero-shot Generalization Comparison}
The learned context-management behavior also transfers beyond BrowseComp-Plus
when the trained policies are evaluated in the same agent-loop framework
without benchmark-specific further training (Table~\ref{tab:main_results}).
These results therefore measure zero-shot agentic transfer, rather than
bare-model generation performance. On Qwen3-32B-Instruct, \textsc{ECHO}
achieves the best average score of 40.2\%, outperforming GRPO (33.6\%) and
SUPO (34.8\%), with especially large gains on evidence-heavy 16-objective QA
and Frames. Under the MoE backbone, \textsc{ECHO} reaches 30.5\% average score
versus 26.9\% for SUPO. The zero-shot ablation rows follow the same trend as
the BrowseComp-Plus training curves: replacing learned selection with semantic
top-$k$ retrieval lowers the dense-backbone average to 38.2\%, and removing
traceable credit further reduces it to 35.6\%. These results suggest that the
benefits of source-indexed reconstruction and traceable credit routing transfer
across tasks and backbone architectures.

\section{Conclusion}

We presented \textsc{ECHO}, a selective turn-memory framework that stores each
completed turn as a source-indexed memory, reconstructs compact contexts through
learned selection, and reuses the same trace to route positive outcome credit to
the final trajectory segment, reused evidence, memory findings, and selection
actions. Under a controlled agent-loop context-management setting,
\textsc{ECHO} improves over the closest collapsed-history baseline, \textsc{SUPO},
while using fewer turns and lower trajectory volume. Ablations and zero-shot
evaluations further show that learned source selection and traceable credit
routing both contribute to robust transfer. These results suggest that
long-horizon agentic RL benefits not only from shortening history, but from
preserving a source-indexed trace of the history that the policy actually
reuses. Looking ahead, we plan to extend \textsc{ECHO} from final-trace
approximation to all-trace credit routing and train it on real software-engineering
tasks with executable feedback, such as SWE-Bench. We will also compare its
rule-triggered context management with action-based memory operations, including
editing, omission, and folding, to further evaluate its generality.

\bibliography{iclr2025_conference}
\bibliographystyle{iclr2025_conference}

\newpage
\appendix

\begin{algorithm}[h]
\caption{\textsc{ECHO} rollout agent loop}
\label{alg:echo_loop}
\begin{algorithmic}[1]
\Require Task prompt $x$, policy $\pi_\theta$, tools $\mathcal{T}$,
context budget $B$, recent-turn budget $K$, selection cap $S$
\State $C \leftarrow x$
\State $\mathcal{M} \leftarrow \emptyset$ \Comment{persistent source-indexed archive}
\State $\mathcal{S} \leftarrow \emptyset$; $G \leftarrow \emptyset$;
$p \leftarrow \emptyset$; $I_{\mathrm{final}} \leftarrow \emptyset$; $i \leftarrow 0$
\While{rollout is not terminated}
    \State Sample assistant response $a_t \sim \pi_\theta(\cdot \mid C)$
    \State Append trainable tokens of $a_t$ to current segment $G$
    \If{$p \neq \emptyset$}
        \State $(i_p,\alpha_{i_p}) \leftarrow p$
        \State Extract finding $m_{i_p}$ from \texttt{<sum\_last\_turn>} in $a_t$
        \State $\mathcal{M} \leftarrow \mathcal{M} \cup
        \{(i_p,\alpha_{i_p},m_{i_p})\}$
        \State Tag finding tokens in $G$ with source id $i_p$
        \State $p \leftarrow \emptyset$
    \EndIf
    \State Parse tool calls from $a_t$
    \If{no tool call remains}
        \State Mark all trainable tokens in $G$ as the final segment
        \State Append $(G,\mathrm{final}=1)$ to $\mathcal{S}$
        \State \textbf{break}
    \EndIf
    \State Execute tool calls and obtain observations $o_t$
    \If{committing $o_t$ would exceed $B$}
        \State Roll back current $a_t$ from $G$ and discard its pending record
        \State Append $(G,\mathrm{final}=0)$ to $\mathcal{S}$
        \State Generate $a^{\mathrm{sel}}
        \sim \pi_\theta(\cdot \mid c^{\mathrm{sel}}(\mathcal{M}))$
        \State Append trainable tokens of $a^{\mathrm{sel}}$ to the saved segment
        and tag its \texttt{<selection>} span
        \State $I \leftarrow
        \rho_{B,S,K}(\mathrm{Parse}(a^{\mathrm{sel}}),\mathcal{M})$
        \State $I_{\mathrm{final}} \leftarrow I$
        \State $C \leftarrow x \oplus \mathrm{Render}(\mathcal{M}[I])
        \oplus x_{\mathrm{cont}}$
        \State Validate $|C|\le B$; otherwise mark the rollout overlong
        \State $G \leftarrow \emptyset$
    \Else
        \State Tag action tokens of $a_t$ in $G$ with source id $i$
        \State $C \leftarrow C \oplus a_t \oplus o_t \oplus x_{\mathrm{turn}}$
        \State $p \leftarrow (i,\mathrm{RenderAction}(a_t))$; $i \leftarrow i+1$
    \EndIf
\EndWhile
\State Build credit masks from final-segment flags, $I_{\mathrm{final}}$,
source ids, and selection spans
\State \textbf{return} $\mathcal{S}$ with credit-token masks; keep $\mathcal{M}$ intact
\end{algorithmic}
\end{algorithm}

\section{Experimental Details}
\label{app:main-results}
Table~\ref{tab:implementation_settings} lists the dataset, rollout, retrieval,
and context-management settings used in our experiments. The main text reports
the settings most relevant to the controlled comparison, including the shared
32k-token working context, the maximum number of reconstruction rounds, and
the resulting effective interaction budget. The remaining details specify the
retrieval backend, tool limits, asynchronous rollout pipeline, and
method-specific memory-selection budgets.

\begin{table}[t]
\centering
\small
\begin{tabular}{ll}
\toprule
Setting & Value \\
\midrule
\multicolumn{2}{l}{\textit{BrowseComp-Plus dataset}} \\
Training examples & 747 \\
Held-out examples & 83 \\
Verifier & Deepseek V4-Flash LLM judge \\

\midrule
\multicolumn{2}{l}{\textit{Tool environment and retrieval}} \\
Tools & \texttt{search}, \texttt{open\_page}, \texttt{finish} \\
Search backend & Qwen3-Embedding-8B retrieval \\
Search return size & Top-5 documents \\
Search snippet length & 2048 chars \\
\texttt{open\_page} limit & 16,000 chars \\
Parallel tool calls & At most 5 per assistant turn \\

\midrule
\multicolumn{2}{l}{\textit{Training and context management}} \\
Rollout group size & 8 \\
Working context budget & 32k tokens \\
Reconstruction rounds & At most 5 \\
Effective interaction budget & Up to 192k tokens \\
Rollout pipeline & Fully asynchronous \\
Hardware & 4 nodes, 8 GPUs per node \\
Partial-rollout staleness & 0.5 \\
Latest turns retained $K$ & 3 \\
Selected historical turns $S$ & At most 8 \\
\bottomrule
\end{tabular}
\caption{
Implementation settings for BrowseComp-Plus training and evaluation.
The effective interaction budget counts the maximum cumulative history
covered across reconstruction rounds; each individual policy call is still
bounded by the 32k-token working context.
}
\label{tab:implementation_settings}
\end{table}

\subsection{Evaluation Benchmarks}
\label{app:benchmarks}
In this section, we provide detailed descriptions of the evaluation protocols and dataset characteristics for the zero-shot generalization experiments. 

\paragraph{Multi-Objective QA.} 
To evaluate the agent's long-range reasoning and context management capabilities under varying cognitive loads, we construct a multi-objective QA benchmark \citep{zhang2025memory}. This benchmark challenges the agent to resolve multiple independent sub-questions within a single extended rollout to synthesize a final answer. We synthesize these multi-objective queries using seed instances from HotpotQA \citep{yang2018hotpotqa}, 2WikiMultihopQA \citep{ho2020constructing}, Bamboogle \citep{press2023measuring}, and Musique \citep{trivedi2022musique}. We scale the task difficulty by aggregating questions to form test sets requiring 2, 4, 8, and up to 16 distinct objectives. Performance is measured using an LLM-as-a-Judge \citep{zheng2023judging} protocol. The evaluator assesses the semantic consistency between the agent's final answer and the ground truth, computing the average success rate across all sub-objectives within a given multi-objective prompt.

\paragraph{Code Generation.}
We evaluate interactive programming capabilities using CodeGym
\citep{du2025codegym} and LoCoBench-Agent \citep{qiu2025locobench}. CodeGym
frames tasks as synthetic interactive environments where agents invoke
problem-specific APIs, together with \texttt{observe()} and \texttt{done()},
rather than writing raw code. Since many CodeGym tasks are either too easy or
too difficult for meaningful comparison, we construct a medium-difficulty
subset using the original \texttt{Qwen3-32B-Instruct} policy before agentic RL
training. Specifically, we first run the base policy four times per instance,
retain instances solved exactly once, and then randomly sample 128 examples
from this candidate pool. This filtering is performed once before evaluating
any trained method, and the same subset is used for all methods. For authentic
software-engineering scenarios, we further evaluate on 128 Python instances
from LoCoBench-Agent. These tasks operate on real codebases of up to 1 million
tokens and require agents to manage cross-file dependencies, investigate bugs,
and perform architectural refactoring with tools for file operations and
semantic search.

\paragraph{Deep Information Seeking.} 
Deep information seeking benchmarks measure the agent's proficiency in orchestrating complex reasoning, multi-step tool use, and long-horizon fact aggregation. We evaluate on GAIA \citep{mialon2024gaia}, which reflects real-world assistant tasks demanding rigorous execution and robust fact-checking. To test the limits of domain expertise, we use Humanity's Last Exam (HLE) \citep{phan2025humanity}, a frontier-level benchmark featuring expert-vetted questions that require deep reasoning rather than surface-level internet retrieval. Finally, we incorporate Frames \citep{krishna2025fact} to assess the agent's capacity to resolve multi-hop queries, integrate conflicting evidence, and maintain critical factual anchors across extensive retrieval trajectories without experiencing history collapse.

\section{Prompt Templates}
  \label{app:prompt-details}

  We provide the prompt templates used by \textsc{ECHO}. Braces denote
  runtime-filled slots. The base BrowseComp-Plus prompt follows the same
  Hermes-style tool-call protocol as the SUPO baseline; \textsc{ECHO} adds the
  turn-summary and selection prompts below.

  \subsection{Base BrowseComp-Plus Prompt}

  \begin{promptpanel}{Tool-Agent Prompt}
  You are an expert research agent tasked with answering the given question
  through iterative tool use.

  \vspace{0.5em}
  \textbf{Question:}\quad \texttt{\{question\}}

  \vspace{0.5em}
  \textbf{Available tools:}
  \begin{itemize}[leftmargin=1.5em, itemsep=1pt, topsep=2pt]
      \item \texttt{search(query)}: retrieve candidate documents.
      \item \texttt{open\_page(docid)}: inspect a retrieved document.
      \item \texttt{finish(answer)}: submit the final answer.
  \end{itemize}

  \textbf{Follow this protocol:}
  \begin{enumerate}[leftmargin=1.5em, itemsep=1pt, topsep=2pt]
      \item Decompose the question into searchable facts.
      \item Search for evidence and open useful pages.
      \item Track constraints, aliases, dates, and sources.
      \item When confident, call \texttt{finish} with a concise final answer.
  \end{enumerate}

  Tool calls must use the Hermes-style format:
  \begin{innercode}
  \textless tool\_call\textgreater\\
  \{\char34 name\char34: \char34...\char34,
  \char34 arguments\char34: \{...\}\}\\
  \textless/tool\_call\textgreater
  \end{innercode}
  \end{promptpanel}

  \subsection{ECHO-Specific Prompts and Hints}

  \begin{promptpanel}{ECHO Extra System Prompt}
  After receiving a tool/function response, your next assistant message must
  include exactly one \texttt{<sum\_last\_turn>...</sum\_last\_turn>} block before
  any new tool/function call.

  The block should be one concise factual sentence summarizing only the latest
  tool/function result.

  After the block, continue in the same assistant message with the next
  tool/function call, or call \texttt{finish} if the final answer is ready. Do not
  stop after the summary.
  \end{promptpanel}

  \begin{promptpanel}{Sum Last Turn Hint}
  Briefly record the latest tool result in
  \texttt{<sum\_last\_turn>...</sum\_last\_turn>}, then continue with the next
  action.
  \end{promptpanel}

  \begin{promptpanel}{Selection Prompt}
  \textbf{System:}\\
  Your operational context is full. Select prior interaction turns that are
  necessary to continue solving the task.

  \vspace{0.5em}
  \textbf{User:}\\
  \textbf{Valid turns:}\\
  \texttt{\{turn\_list\}}

  \vspace{0.5em}
  \textbf{Rules:}
  \begin{itemize}[leftmargin=1.5em, itemsep=1pt, topsep=2pt]
      \item Output exactly one \texttt{<selection>...</selection>} block.
      \item Do not call any function/tool in this turn.
      \item Do not answer the original task.
      \item Do not include \texttt{<think>}, tool calls, markdown fences, or text
      outside the selection tags.
      \item Select only turns that contain reusable evidence, constraints, failed
      attempts, or the next planned action.
      \item Prefer older turns that are not already covered by automatically
      retained recent turns.
      \item If no older turn is necessary, return an empty
      \texttt{<selection></selection>} block.
  \end{itemize}

  \vspace{0.5em}
  \textbf{Output format:}
  \begin{innercode}
  \textless selection\textgreater
  [zero or more lines, each formatted as turn\_N: reason]
  \textless/selection\textgreater
  \end{innercode}
  \end{promptpanel}

  \begin{promptpanel}{Selection Hint}
  Valid selection indices: \texttt{turn\_0} through
  \texttt{turn\_\{n\_hist-1\}}.

  Select at most \texttt{\{selection\_max\_turns\}} historical turns; output an
  empty selection if the automatically retained recent turns are enough.

  The latest turns \texttt{turn\_\{recent\_start\}} through
  \texttt{turn\_\{n\_hist-1\}} will be retained automatically; select any
  additional older turns that are still needed.

  \vspace{0.5em}
  In our BrowseComp-Plus experiments, \texttt{selection\_max\_turns = 8} and
  \texttt{echo\_recent\_turns = 3}.
  \end{promptpanel}

\end{document}

%% file: math_commands.tex
\usepackage{amsmath,amsfonts,bm}

\def\eqref#1{equation~\ref{#1}}

\def\1{\bm{1}}

\DeclareMathAlphabet{\mathsfit}{\encodingdefault}{\sfdefault}{m}{sl}
\SetMathAlphabet{\mathsfit}{bold}{\encodingdefault}{\sfdefault}{bx}{n}

